\documentclass[conference]{IEEEtran}
\IEEEoverridecommandlockouts
\usepackage{cite}
\usepackage{amsmath,amssymb,amsfonts}
\usepackage{algorithmic}
\usepackage{multirow}
\usepackage{makecell}
\usepackage{graphicx}
\usepackage{textcomp}
\usepackage{xcolor}
\def\BibTeX{{\rm B\kern-.05em{\sc i\kern-.025em b}\kern-.08em
    T\kern-.1667em\lower.7ex\hbox{E}\kern-.125emX}}
\begin{document}
\title{IDVT: Interest-aware Denoising and View-guided Tuning for Social Recommendation}

\author{\IEEEauthorblockN{Dezhao Yang}
\IEEEauthorblockA{\textit{School of Computer Science and Technology } \\
\textit{Harbin Institute of Technology}\\
Shenzhen, China \\
yangdezhaoomg@gmail.com}
\and
\IEEEauthorblockN{Jianghong Ma}
\IEEEauthorblockA{\textit{School of Computer Science and Technology} \\
\textit{Harbin Institute of Technology}\\
Shenzhen, China \\
majianghong@hit.edu.cn}
\and
\IEEEauthorblockN{Shanshan Feng}
\IEEEauthorblockA{\textit{Centre for Frontier AI Research} \\
\textit{Agency for Science} \\
\textit{Technology and Research} \\
Singapore \\
victor\_fengss@foxmail.com}
\and
\IEEEauthorblockN{Haijun Zhang}
\IEEEauthorblockA{\textit{School of Computer Science and Technology} \\
\textit{Harbin Institute of Technology}\\
Shenzhen, China \\
hjzhang@hit.edu.cn}
\and
\IEEEauthorblockN{Zhao Zhang}
\IEEEauthorblockA{\textit{School of Computer Science and Information Engineering} \\
\textit{Hefei University of Technology}\\
Hefei, China \\
cszzhang@gmail.com}
}

\maketitle

\begin{abstract}
In the information age, recommendation systems are vital for efficiently filtering information and identifying user preferences. Online social platforms have enriched these systems by providing valuable auxiliary information. Socially connected users are assumed to share similar preferences, enhancing recommendation accuracy and addressing cold start issues. However, our statistical analysis indicates a significant amount of noise in the social network, where many socially connected users do not share common interests. We believe that incorporating low-quality social connections into the recommendation process will harm the performance. To address this issue, we propose an innovative \underline{I}nterest-aware \underline{D}enoising and \underline{V}iew-guided \underline{T}uning (IDVT) method for the social recommendation. The first ID part effectively denoises social connections. Specifically, the denoising process considers both social network structure and user interaction interests in a global view. Moreover, in this global view, we also integrate denoised social information (social domain) into the propagation of the user-item interactions (collaborative domain) and aggregate user representations from two domains using a gating mechanism. To tackle potential user interest loss and enhance model robustness within the global view, our second VT part introduces two additional views (local view and dropout view) with different generation strategies for fine-tuning user representations in the global view through contrastive learning. Extensive evaluations on real-world datasets with varying noise ratios demonstrate the superiority of IDVT over state-of-the-art social recommendation methods. The codes are released at: https://anonymous.4open.science/r/IDVT-main-C192.
\end{abstract}


\section{Introduction}
With the rapid growth of multimedia platforms such as Twitter, Instagram, and TikTok, recommendation systems have emerged as indispensable tools for filtering and discerning user preferences effectively \cite{ricci2010introduction}. However, relying solely on user-item interactions may face the issue of data sparsity, making it necessary to introduce auxiliary information to enhance the effectiveness of recommendation systems \cite{bayer2017generic}. Social networks exert a significant influence on users during their usage of multimedia platforms, owing to the effects of social homophily and social influence ~\cite{fu2021dual, han2022dh, yang2022large}. Furthermore, contrastive learning (CL) methods have gained prominence as a prevailing strategy to mitigate data sparsity concerns \cite{wang2023unbiased,yu2023contrastive}. However, existing social recommendation algorithms and CL algorithms face two major challenges as follows:

First, it is widely recognized that users with social connections often exhibit similar interaction preferences~\cite{mcpherson2001birds,cialdini2004social}. Leveraging these social connections not only improves recommendation accuracy but also effectively addresses the cold start problem.  However, it should be noted that not all social connections entail common interactions, leading to what can be referred to as ``noises" in real-life social networks. Real-life social networks are indeed susceptible to a significant amount of noise. To substantiate this observation, we introduce a novel metric, termed the Noise Ratio, aimed at quantifying the noise present within social networks. A detailed analysis of the Noise Ratio is presented in Section III.A of our study. We believe that incorporating low-quality social connections into the recommendation process may compromise the overall system performance. However, most Graph Neural Networks (GNNs) based social recommendation models overlook the presence of noises in social networks. Among the few that consider social noises, DESIGN~\cite{tao2022revisiting} employs knowledge distillation, which can be inefficient as student models frequently struggle to adequately learn the knowledge from the teacher models. While DSL~\cite{wang2023denoised} fails to capitalize on both collaborative and social domains for the recommendation task, where shared user interests in the collaborative domain could potentially guide the denoising process. SHaRe~\cite{jiang2024challenging} only considers user interests in the graph rewiring process, disregarding the structural information of the social network. 

Second, typically, CL adopts a view generation approach to produce two distinct views from the original data, which are then optimized using a contrastive loss. View generation strategies are generally classified into two types: those with augmentation and those without augmentation strategies. The former strategies introduce the randomness that helps to learn noise-invariant representations, while the latter strategies preserve the semantic information of the original data. However, these strategies are never integrated into recommendation methods. Given their potential benefits, one might speculate on the applicability of these strategies in mitigating the challenges faced by social recommendation systems.

To address the dual challenges outlined above, we propose a novel model named  \underline{I}nterest-aware \underline{D}enoising and \underline{V}iew-guided \underline{T}uning method (IDVT) for Social Recommendation. The IDVT model consists of three modules, with the primary module in a global view. Within this module, three components are employed: an interest-aware denoising component, which mitigates noisy social information in real-world social networks; a cross-domain gated aggregation component, integrating denoised social information into the propagation of user-item interactions; a social integrated propagation component, which aggregates user representations from interaction and social domains through a gating mechanism. To complement these efforts, two ancillary modules are implemented: the local view and the dropout view. The former aims to learn semantic-preserving representations, while the latter focuses on learning noise-invariant representations. In summary, these two auxiliary modules collectively form a novel two-level view-guided tuning process. Specifically, the strategies proposed in IDVT for addressing the two challenges are as follows:

First, the interest-aware denoising component in the primary module introduces a strategy that identifies and extracts valuable social connections characterized by shared user interests. This strategy incorporates an innovative scoring mechanism that evaluates the reliability of social connections between users by taking into account both their interests and the underlying structure of the social network. Subsequently, a predetermined threshold is applied to filter out noisy user-user edges, thereby enhancing the quality of the retained connections.

Second, in the primary module, two limitations still persist. One concerns the potential loss of original user interests during the learning process of user representations across different domains. The other limitation pertains to the granularity of denoising social connections, which may be considered coarse when operating below a predetermined threshold. To address these limitations, we introduce two additional auxiliary modules aimed at refining user representations in the global view through CL. The first module, referred to as the local view, tackles the first limitation by using a view generation strategy without augmentation, thereby focusing on interest-aware tuning. It achieves this by leveraging information from the interaction graph to emphasize user interests. The second module, termed the dropout view, addresses the second limitation by using a view generation strategy with augmentation. It enhances data-enriched tuning by utilizing edge dropout to augment data and enhance the robustness of the denoising process.

Our contributions can be summarized as follows:
\begin{itemize}
\item We propose a novel social denoising method in the global view that considers both user interests and the underlying social graph structure. Subsequently, this denoised social information is integrated with interaction data using a cross-domain propagation and aggregation approach, enabling the effective learning of user and item representations across both social and collaborative domains.

\item We introduce two novel additional views to refine global view representations via CL. The local view emphasizes interest-aware tuning to mitigate potential user interest loss in the denoising module without using augmentation strategies. The dropout view utilizes augmentation strategies to enhance overall robustness through data-enriched tuning.

\item We evaluate the performance of IDVT on three real-world datasets with varying noise ratios in the social graphs. The experimental results reveal the superiority of our proposed model over state-of-the-art methods.

\end{itemize}

\section{Related Work}
\subsection{Social Recommendation}
With the rise of social media platforms, social networks are often used as auxiliary information to enhance recommendation performance. Due to the superior capabilities of Graph Neural Network (GNN) technology in learning representations of graph-structured data \cite{wang2019neural,wang2020multi,ji2020dual,wang2020disentangled,prakash2023enhancing}, GNN-based social recommendation algorithms have become mainstream. Moreover, many algorithms have been designed with self-supervised tasks to enhance model performance. We categorize social recommendation models into GNN-based and Self-Supervised Learning (SSL) based methods. In the GNN-based category, GraphRec~\cite{fan2019graph} utilizes attention mechanisms to effectively capture interactions and social connections. DiffNet~\cite{wu2019neural} and DiffNet++~\cite{wu2020diffnet++} both introduce a layer-wise influence propagation structure for recursive social diffusion. DESIGN~\cite{tao2022revisiting} incorporates knowledge distillation to address social network noise. Among SSL-based methods,  ESRF~\cite{yu2020enhancing} employs a deep adversarial framework to enhance performance. MHCN~\cite{yu2021self} uses multiple motifs to explore semantic information within graph structures, while SEPT~\cite{yu2021socially} proposes a socially-aware tri-training framework empowered by SSL. DcRec~\cite{wu2022disentangled} presents a novel approach in the form of a disentangled CL framework. DSL~\cite{wang2023denoised} leverages cross-view information to denoise social networks.  SHaRe~\cite{jiang2024challenging} rewires social graph by cutting unreliable edges and adding highly homophilic edges, and samples highly homophilic user pairs for constructing contrastive learning task.

Our approach diverges the above studies in two key aspects. \textbf{(1)} Unlike DESIGN, DSL, and SHaRe, we incorporate both user interests and social network structures during the denoising process. \textbf{(2)} We utilize a cross-domain propagation approach to learn user and item embeddings, whereas existing methods typically focus on either the social or interaction domain for this purpose.

\subsection{Contrastive Leaning for Recommendation}
Amidst the data sparsity challenges in recommendation systems, SSL has emerged as a promising solution for extracting information from unlabelled data~\cite{yu2023self,huang2022self}. One popular SSL method is CL, which maximizes the agreement between different views~\cite{lin2022improving}. Thus, the generation of diverse views plays a crucial role in CL. View generation strategies can be broadly categorized into two types: those with augmentation and those without augmentation strategies. For augmentation strategies, SGL~\cite{wu2021self} employs data augmentation techniques such as node dropout, edge dropout, and random walking on the interaction graph to create different views. SimGCL~\cite{yu2022graph} introduces uniform noises to node representations for view generation. LightGCL~\cite{cai2023lightgcl} uses Single Value Decomposition (SVD) on the interaction graph to distill valuable information as the second view. RGCF \cite{tian2022learning} considers noise in the user-item interaction graph. RGCF first denoises the original interaction graph to generate a denoised view, then add reliable edges to generate a diverse view, preserving recommendation diversity while removing noise. For strategies without augmentation, MIDGN~\cite{zhao2022multi} showcases the benefits of user preferences and item associations through CL from both interaction and bundle views. S-MBRec~\cite{gu2022self} construct views through target and auxiliary behaviors. CGCL \cite{he2023candidate} considers the semantic relationships between users and candidate items across different embedding layers.

In contrast to the above investigations that typically rely on a single view generation strategy for constructing CL tasks, IDVT employs two different view generation strategies to fine-tune representations in the primary module.

\section{Preliminary}
\subsection{Data Observation}
\begin{figure}[t]
\centering
\includegraphics[width=1\columnwidth]{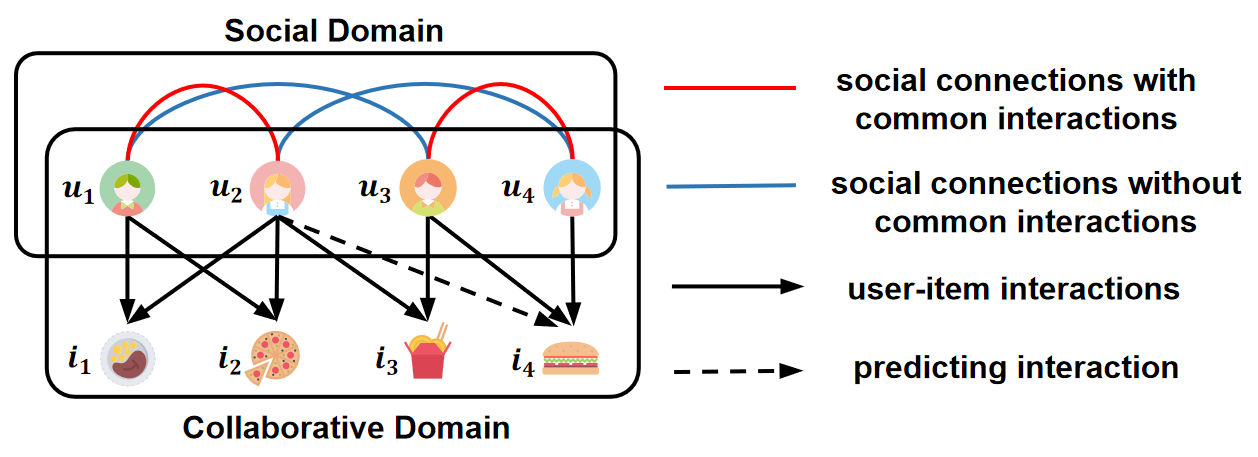} 
\caption{Illustrative example of Noise Ratio. }
\label{fig1}
\end{figure}

In our study, we identify social connections lacking common interactions as ``noises."  For instance, in Fig. \ref{fig1}, $u_2$ and $u_4$ are socially connected despite not sharing interests. This can lead to inappropriate recommendations, such as recommending $u_2$ with $u_4$'s favorite products, which is clearly inappropriate given their lack of common interests. To quantify these noises, we introduce a novel metric called Noise Ratio, which represents the proportion of social connections in the social network that lack common interactions. For example, the noise ratio in Fig.\ref{fig1} is 50\%, as evidenced by 2 social links without common interactions among the four total social links. We conduct a statistical analysis of the Noise Ratio in three real-world social network datasets, respectively Flickr, Ciao, and Yelp. The results are presented in Table \ref{tab:statistic1}. The computed noise ratios for Flickr, Ciao, and Yelp, are \textbf{81\%}, \textbf{45\%}, and \textbf{29\%}, respectively, indicating a substantial number of socially connected users without common interactions. These socially connected but interactionally distant pairs ought to be eliminated, as they do not effectively capture shared user interests.

\begin{table}[htbp]
\centering
\caption{The statistics of Noise Ratio.}
\begin{tabular}{c|c|c|c}
\hline
Dataset         & Flickr   & Ciao     & Yelp    \\ \hline
\hline
Number of Social Connections  & 117,458   & 95,995    & 88,582   \\ \hline
\makecell{Number of Social Connections \\ without Common Interactions} & 95,140   & 43,197   &25,688    \\ \hline
Noise Ratio         & 81\%     & 45\%      & 29\%     \\ 
\hline
\end{tabular}
\vspace{-3mm}
\label{tab:statistic1} 
\end{table}

\subsection{Problem Formulation}
In social recommendation, the sets of users and items are denoted as $\mathcal{U}=\left\{u_1, \ldots, u_n\right\}$ and $\mathcal{I}=\left\{i_1, \ldots, i_m\right\}$, where $n$ and $m$ represent the number of users and items, respectively. Two essential graphs are involved: the interaction graph $\mathcal{G}r$ and the social graph $\mathcal{G}s$. The interaction matrix is denoted by $\mathbf{R} \in \mathbb{R}^{n \times m}$, where $r_{ui}=1$ indicates the existence of interaction between user $u$ and item $i$, $r_{ui}=0$ otherwise. The social matrix is denoted by $\mathbf{S} \in \mathbb{R}^{n \times n}$, where $s_{uv}=1$ signifies the directed social relation from user $u$ to user $v$, $s_{uv}=0$ otherwise. We use social network $\mathbf{S} $ to enhance the performance of top-$K$ tasks in recommender systems, which involve recommending a list of $K$ items to users with whom they have not interacted before.

The main goal of this research is to design a recommendation system that can effectively alleviate the noise present within social networks and leverage diverse views to enhance user representations for improving final predictions.

\section{Methodology}


\subsection{Overview}
\begin{figure*}[t]
\vspace{-3mm}
\centering
\includegraphics[width=1\linewidth]{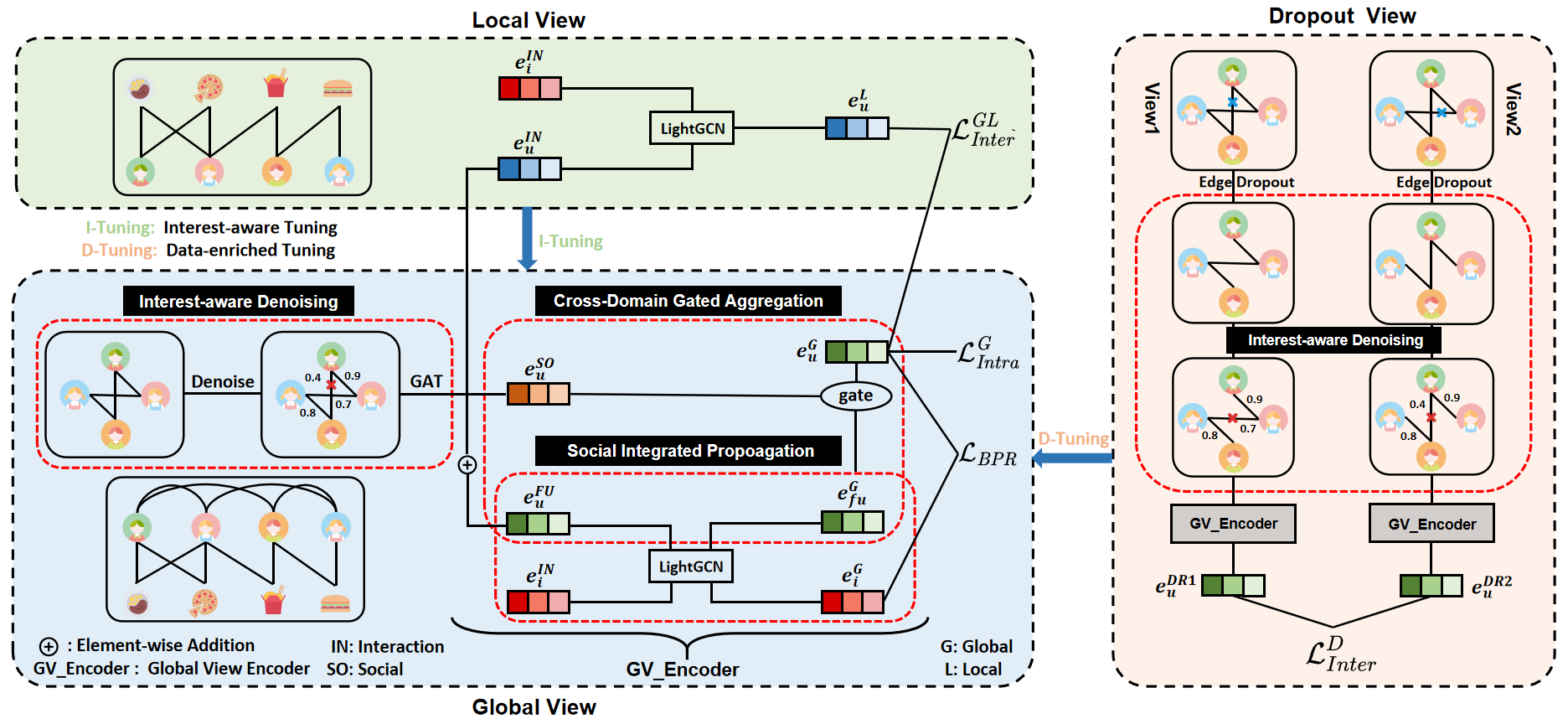} 
\vspace{-5mm}
\caption{Illustration of the proposed IDVT model. The primary module learns user representations in the ``Global View". The ancillary modules in the ``Local View" and the ``Dropout View" refine user representations in the ``Global View". While the former module accentuates interest-aware fine-tuning, the latter module is dedicated to data-enriched fine-tuning.} 
\label{fig:framework} 
\end{figure*}

Fig.\ref{fig:framework} illustrates the tripartite structure of the proposed IDVT model, featuring three essential modules. The primary module is situated in the ``Global View". Accompanying this, the ``Local View" and the ``Dropout View", serving as auxiliary modules, contribute to the refinement of user representations in the ``Global View". While the former auxiliary module focuses on interest-aware fine-tuning, the latter auxiliary module pursues data-enriched fine-tuning, creating a comprehensive two-level view-guided tuning process.





\subsection{Global View}
In this section, we present the approach to learning user representations by leveraging both the social and interaction graphs within the global view. 

\subsubsection{Interest-aware Denoising.} 
In the recommendation scenario, it is observed that many users with social connections do not share common interests, resulting in limited interactions with common items. To address this challenge and enhance the removal of redundant edges in the social graph, we consider the graph structure of the social network and the individual interests of users. Our approach draws inspiration from  RGCF ~\cite{tian2022learning}, which is a pioneering denoising method tailored to GNN-based recommendation systems, with a specific focus on denoising the interaction graph. By incorporating similar denoising principles in our IDVT model, we aim to further improve the reliability of edge removal in the social graph, as captured by the following formula: 
\begin{figure}[t]
\centering
\includegraphics[width=0.9\columnwidth]{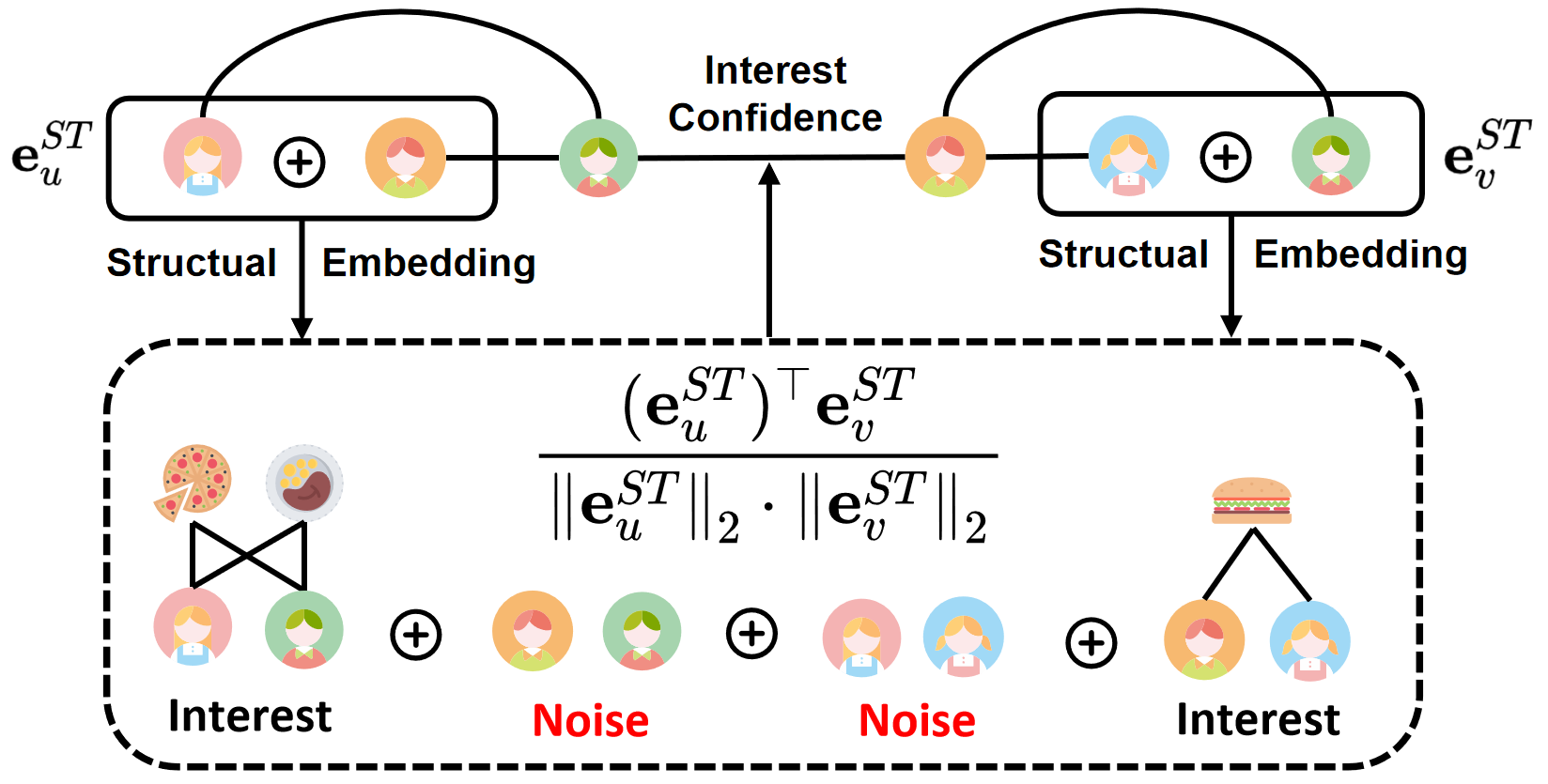}
\vspace{-3mm}
\caption{Illustrative example for IC of user $u$ and $v$. }
\vspace{-5mm}
\label{fig:ID} 
\end{figure}
\begin{equation}
\mathbf{E}_U^{ST}=\mathbf{SE}_U^{IN},
\end{equation}
where $\mathbf{E}_U^{IN} \in \mathbb{R}^{ n \times d}$ represents the user embedding matrix learned from the one-hot encoding of user Ids, while $\mathbf{E}_U^{ST} \in \mathbb{R}^{ n \times d}$ denotes the user structural embedding matrix. Notably, instead of directly utilizing one-hot encoding for users, we extract users' one-hop social neighbors, which encapsulate structural information from the social graph. 

To quantify the reliability degree of the social connection between user $u$ and user $v$, we utilize the corresponding embeddings $\mathbf{e}_u^{ST}$ and $\mathbf{e}_v^{ST}$ from the user structural embedding matrix $\mathbf{E}_U^{ST}$. To achieve this, we employ the cosine similarity function, as shown in Eq.$\left(2\right)$, which calculates the cosine similarity between the two embeddings. To ensure non-negativity and enhance the interpretability and effectiveness of the reliability measurement, we then normalize the computed cosine similarity using Eq.$\left(3\right)$. 
\begin{equation}
\cos \left(\mathbf{e}_u^{ST}, \mathbf{e}_v^{ST}\right)=\frac{(\mathbf{e}_u^{ST})^{\top} \mathbf{e}_v^{ST}}{\left\|\mathbf{e}_u^{ST}\right\|_2 \cdot\left\|\mathbf{e}_v^{ST}\right\|_2}, 
\end{equation}
\begin{equation}
IC_{u, v}=\left(\cos \left(\mathbf{e}_u^{ST}, \mathbf{e}_v^{ST}\right)+1\right) / 2,
\end{equation}
where $IC_{u, v}$ represents the Interest Confidence (IC) score, which quantifies the reliability of the social connection between user $u$ and user $v$. Fig.\ref{fig:ID} visually demonstrates the process, depicting the computed IC value divided into four segments, of which two segments are noises, while the remaining two segments are shared interests. Therefore, the interests shared between their respective neighbors are considered. 

To address the potential noise introduced by the cosine similarity calculation, we apply a specified threshold to eliminate noisy edges in the social graph $\mathcal{G}_s$. This process is denoted as
\begin{equation}
\widetilde{\mathcal{G}_s}=\operatorname{Denoise}\left(\mathcal{G}_s,\text{Threshold}\right),
\end{equation}
where edges are selectively removed based on the reliability degree (represented by the $IC$ value) of each edge. Those edges with a reliability degree lower than the threshold are identified as noisy connections and, consequently, eliminated from the social graph. The aim of this interest-aware denoising operation is to retain solely those social connections that exhibit sufficiently high reliability, effectively enhancing the robustness and informative quality of the social graph. Subsequently, we employ the GAT to learn the social domain embedding as follows: 
\begin{equation}
\mathbf{E}_U^{SO}=\operatorname{GAT}\left(\mathbf{E}_U^{IN},\widetilde{\mathcal{G}_s}\right),
\end{equation}
where $\mathbf{E}_U^{SO} \in \mathbb{R}^{n \times d}$ inherently captures the social relationships between users in the context of social influence. The GAT used here aids the extraction of meaningful social representations from the denoised social graph.

\subsubsection{Social Integrated Propagation.}
In our study, we utilize the LightGCN model, denoted as LGCN, for the interaction graph encoder due to its simplicity and efficacy in capturing user-item interactions. The implementation details of the encoder, as shown in Eq.$\left(6\right)$ below, demonstrate how LGCN is applied to the interaction graph:
\begin{equation}
\begin{aligned}
\mathbf{e}_u^{(k+1)}=\sum_{i \in \mathcal{N}_u} \frac{1}{\sqrt{\left|\mathcal{N}_u\right|} \sqrt{\left|\mathcal{N}_i\right|}} \mathbf{e}_i^{(k)} \\
\mathbf{e}_i^{(k+1)}=\sum_{u \in \mathcal{N}_i} \frac{1}{\sqrt{\left|\mathcal{N}_i\right|} \sqrt{\left|\mathcal{N}_u\right|}} \mathbf{e}_u^{(k)}
\end{aligned}.
\end{equation}

Different from general social recommendation models that follow a disentangled learning paradigm~\cite{sharma2022survey}, our proposed model adopts a novel strategy by directly integrating social information into the propagation of the user-item interaction graph. This integration jointly considers both users' social connections and individual interests when learning the user and item embeddings. Specifically, the user and item embeddings can be represented as
\begin{equation}
\mathbf{E}_U^{FU}=\mathbf{E}_U^{SO} \oplus \mathbf{E}_U^{IN},
\end{equation}
\begin{equation}
\mathbf{E}_{FU}^{G},\mathbf{E}_I^{G}=\operatorname{LGCN}\left(\mathbf{E}_U^{FU},\mathbf{E}_I^{IN},{\mathcal{G}_r}\right),
\end{equation}
where $\oplus$ denotes element-wise addition. $\mathbf{E}_U^{FU} \in \mathbb{R}^{ n \times d}$ represents the fused user embedding matrix, which is learned through fusing information from both social and interaction graphs. This fusion process enables the model to effectively capture the intricate dynamics of social connections and user preferences. Additionally, $\mathbf{E}_{FU}^{G} \in \mathbb{R}^{ n \times d}$ and $\mathbf{E}_I^{G}\in \mathbb{R}^{ m \times d}$
serve as the fused user global embedding matrix and item global embedding matrix, respectively. They capture the integrated global representations of users and items. 

\subsubsection{Cross-Domain Gated Aggregation.}
While we integrate social information into the interaction graph, it is noteworthy that $\mathbf{E}_{FU}^{G}$ predominantly emphasizes user interest, given its learning process through the user-item interaction graph. Nevertheless, we also recognize the considerable importance of denoised social embeddings in enhancing our model's overall performance. To effectively aggregate the representations from both domains, we employ a gating mechanism~\cite{wu2022eagcn}, which assumes a pivotal role in the integration process. This gating mechanism enables us to carefully balance the contributions of the user interest representation and the denoised social representation, thus allowing our model to capitalize on the respective strengths of both domains. The aggregation process is stated below:
\begin{equation}
\mathbf{E}_U^{G}=\operatorname{Gate}\left(\mathbf{E}_{FU}^{G},\mathbf{E}_U^{SO}\right),
\end{equation}
where the details and operation of the gating mechanism are described as follows:
\begin{equation}
\begin{aligned}
\mathbf{g}_u=\sigma\left(\mathbf{W}_G^1 \  \mathbf{e}_{fu}^{G}+\mathbf{W}_G^2 \ \mathbf{e}_u^{SO}\right) \\
\mathbf{e}_u^{G}=\mathbf{g}_u \odot \mathbf{e}_{fu}^{G} +\left(1-\mathbf{g}_u\right) \odot \mathbf{e}_u^{SO}
\end{aligned},
\end{equation}
where $\odot$ stands for element-wise product, $\sigma \left(\cdot\right)$ denotes sigmoid function, $\mathbf{W}_G^1, \mathbf{W}_G^2 \in \mathbb{R}^{ d \times d} $ are transform matrices that dynamically control the weights of the two domains.

\vspace{-1mm}
\subsection{Local View}

In this section, we achieve interest-aware tuning by refining user representations in the global view using information from the local view created without augmentation, thereby adapting user embeddings based on individual interests and preferences.

Inspired by the multi-level cross-view contrastive framework for knowledge-aware recommendation~\cite{zou2022multi}, we explore a multi-view approach to enhance node representations in our model. Notably, in our approach, the embeddings learned from the local view serve as moderators and do not participate in recommendation predictions. In the global view, we learn user and item embeddings from both the social and interaction graphs, considering the broader social influences on user preferences. However, we also recognize the significance of directly capturing user interests from the interaction graph, which provides a more detailed and personalized understanding of individual user preferences. To achieve this, we apply contrastive learning to fine-tune the global user embeddings, effectively integrating the global and local views. This process allows us to refine and tailor the user representations to their specific interests. 

To obtain the user local embedding matrix, we utilize the same GNN encoder used in the global view as follows:
\begin{equation}
\mathbf{E}_{U}^{L} =\operatorname{LGCN}\left(\mathbf{E}_U^{IN},\mathbf{E}_I^{IN},{\mathcal{G}_r}\right).
\end{equation}

The local view complements the global view by involving contrastive learning. To enhance the discriminative power of user embeddings across different views, 
we design a contrastive loss that comprises two components: inter-view and intra-view contrastive learning .

In inter-view contrastive learning, we aim to maximize the similarity between user embeddings from different views that correspond to the same user. This encourages the model to capture common patterns and preferences exhibited by the user across various perspectives. Specifically, considering a triple pair $\{\mathbf{e}_{u}^{G}, \mathbf{e}_{u}^{L}, \mathbf{e}_{v}^{L}\}$, we regard $\{\mathbf{e}_{u}^{G}, \mathbf{e}_{u}^{L}\}$ as a positive pair since they represent the same user in different views. Conversely, $\{\mathbf{e}_{u}^{G}, \mathbf{e}_{v}^{L}\}$ are considered as negative pairs. The inter-view contrastive loss is formulated as follows:
\begin{equation}
\mathcal{L}_{Inter}^{GL}=\sum_{u \in \mathcal{U}}-\log \frac{\exp \left(\left(\mathbf{e}_{u}^{G} \cdot \mathbf{e}_{u}^{L} / \tau\right)\right)}{\sum_{v \in \mathcal{U}} \exp \left(\left(\mathbf{e}_{u}^{G} \cdot \mathbf{e}_{v}^{L} / \tau\right)\right)},
\end{equation}
where InfoNCE ~\cite{gutmann2010noise} is adopted in the loss function, and $\tau$ is the temperature parameter.

In contrast, intra-view contrastive learning focuses on maximizing the similarity between positive pairs of user embeddings within the same view. This approach enhances the model's ability to discern the intricate nuances and subtle patterns of user embeddings within the global view. Specifically, we consider the same user as positive pairs and different users as negative pairs from the same view. The intra-view contrastive loss is expressed as follows:
\begin{equation}
\mathcal{L}_{Intra}^{G}=\sum_{u \in \mathcal{U}}-\log \frac{\exp \left(\left(\mathbf{e}_{u}^{G} \cdot \mathbf{e}_{u}^{G} / \tau\right)\right)}{\sum_{v \in \mathcal{U}} \exp \left(\left(\mathbf{e}_{u}^{G} \cdot \mathbf{e}_{v}^{G} / \tau\right)\right)}.
\end{equation}

By incorporating both inter-view and intra-view contrastive learning, our model can leverage the diverse sources of information from the different views, yielding more discriminative and informative user embeddings. 

\vspace{-1mm}
\subsection{Dropout View}
We further investigate data-enriched tuning to enhance user representations in the global view. This involves refining the user embeddings based on information from the dropout view created with augmentation, where we randomly remove edges from the social graph.

In the Interest-aware Denoising part, the approach of removing edges below a fixed threshold is considered coarse and less robust in addressing redundant edges within the social graph. To overcome this limitation, we introduce a new global view to improve the robustness of the denoising process. In this view, data is enriched by applying edge dropout, a technique that involves randomly removing edges from the social graph based on a specified drop ratio: 
\begin{equation}
\begin{aligned}
\mathcal{G}_s^1 = \operatorname{Edge-Dropout}\left(\mathcal{G}_s,Drop\_Ratio\right) \\
\mathcal{G}_s^2 = \operatorname{Edge-Dropout}\left(\mathcal{G}_s,Drop\_Ratio\right)
\end{aligned},
\end{equation}
where $\mathcal{G}_s^1,\mathcal{G}_s^2$ are two different social graphs obtained by edge dropout with the same dropout ratio. By introducing randomness through edge dropout, we can explore various potential graph configurations and capture diverse patterns of social interactions. This randomization not only improves the robustness of the denoising process but also uncovers hidden connections within the social graph representation. 

Next, we apply the Interest-aware Denoising (ID) and Global View Encoder to obtain dropout view embeddings for each of these social graphs as follows:
\begin{equation}
\begin{aligned}
\mathbf{E}_U^{DR1}= \operatorname{GV\_Encoder}\left( \operatorname{ID}\left(\mathcal{G}_s^1\right)\right) \\
\mathbf{E}_U^{DR2}= \operatorname{GV\_Encoder}\left( \operatorname{ID}\left(\mathcal{G}_s^2\right)\right)
\end{aligned},
\end{equation}
where $\mathbf{E}_U^{DR1} \in \mathbb{R}^{ n \times d}$, $\mathbf{E}_U^{DR2} \in \mathbb{R}^{ n \times d}$ are dropout view user embedding matrices for view $1$ and view $2$, respectively. The GV\_Encoder contains social integrated propagation and cross-domain gated aggregation.

To further refine user embeddings, we employ the contrastive learning approach, aiming to maximize the similarity between embeddings of the same user obtained from different dropout views and minimize the similarity between embeddings of different users. The contrastive loss is defined as 
\begin{equation}
\mathcal{L}_{Inter}^{D}=\sum_{u \in \mathcal{U}}-\log \frac{\exp \left(\left(\mathbf{e}_{u}^{DR1} \cdot \mathbf{e}_{u}^{DR2} / \tau\right)\right)}{\sum_{v \in \mathcal{U}} \exp \left(\left(\mathbf{e}_{u}^{DR1} \cdot \mathbf{e}_{v}^{DR2} / \tau\right)\right)}.
\end{equation}

\subsection{Model Training}
For the main recommendation task, we adopt the Bayesian Personalized Ranking (BPR) loss, a well-designed ranking objective function for recommendation. 
The BPR loss is defined as follows:
\begin{equation}
\mathcal{L}_{BPR}=\sum_{(u, i, j) \in O}-\log \sigma\left(\hat{y}_{u i}-\hat{y}_{u j}\right),
\end{equation}
where $\hat{y}_{u i}=\mathbf{e}_{u}^{G \top}\mathbf{e}_{i}^{G}$ and $O=\left\{(u, i, j) \mid r_{ui}=1, r_{uj}=0\right\}$ is a collection of triples obtained by negative sampling. For each interaction pair $\left(u,i\right)$ in the training set, we randomly select an item that user $u$ has not interacted with as a negative example. 

To jointly optimize the recommendation and contrastive losses, we adopt a multi-task learning strategy:
\begin{equation}
\begin{aligned}
\mathcal{L}=&\mathcal{L}_{BPR} +\lambda_1\left(\beta\mathcal{L}_{Inter}^{GL} + \left( 1-\beta\right)\mathcal{L}_{Intra}^{G}\right) \\
 & +\lambda_2 \mathcal{L}_{Inter}^{D}+\lambda_3\|\Theta\|_2
\end{aligned},
\end{equation}
where $\Theta$ represents the set of learned user and item embeddings, $\beta$ balances the inter- and intra-contrastive ratio, $\lambda_1$ and $\lambda_2$ control the weights of two view-guided tuning modules, and $\lambda_3$ is the regularization coefficient.

\begin{table}[htbp]
\centering
\vspace{-2mm}
\caption{The statistics of datasets.}
\vspace{-3mm}
\begin{tabular}{c|c|c|c}
\hline
    Dataset             & Flickr   & Ciao     & Yelp    \\ \hline
\hline
    Users            & 5,642     & 5,836     & 4,846    \\
    Items            & 21,176    & 10,708    & 5,695    \\
    Rating           & 199,500   & 142,805   & 142,950  \\
  Rating Density     & 0.16\%   & 0.22\%   & 0.51\%  \\
     Relation        & 117,458   & 95,995    & 88,582   \\
  Relation Density   & 0.36\%   & 0.28\%   & 0.32\%  \\

 \hline
\end{tabular}
\label{tab:statistic2} 
\end{table}

\begin{table*}[htbp]
\centering
\caption{Overall Top-5 recommendation performance comparison of different recommender models.}
\vspace{-3mm}
\label{Table:overall}
\renewcommand\arraystretch{1.0}
\begin{center}
\resizebox{0.93\linewidth}{!}{ 
\begin{tabular}{cc|cccccccccc|c}
\hline
Dataset&Metric&BPR&DiffNet&LightGCN&ESRF&SEPT&MHCN&DESIGN&DSL&SHaRe&IDVT&Improv.\\ \hline
\hline

\multirow{4}{*}{Flickr}
&H@5&0.0018&0.0021&0.0048&0.0032&0.0033&0.0045&\underline{0.0053}&0.0051&0.0050&\textbf{0.0062}&16.9\%\\
&P@5&0.0027&0.0033&0.0075&0.0050&0.0051&0.0069&\underline{0.0081}&0.0079&0.0077&\textbf{0.0095}&17.2\%\\		
&R@5&0.0024&0.0029&0.0057&0.0035&0.0045&0.0060&\underline{0.0068}&0.0061&
0.0067&\textbf{0.0083}&22.0\%\\			
&N@5&0.0033&0.0039&0.0090&0.0057&0.0059&0.0082&\underline{0.0098}&0.0092&\underline{0.0098}&\textbf{0.0116}&18.3\%\\

\hline
\multirow{4}{*}{Ciao}
&H@5&0.0250&0.0254&0.0420&0.0415&0.0423&0.0422&0.0420&0.0424&\underline{0.0434}&\textbf{0.0496}&14.2\%\\	
&P@5&0.0281&0.0325&0.0454&0.0449&0.0458&0.0457&0.0455&0.0459&\underline{0.0470}&\textbf{0.0537}&14.2\%\\	
&R@5&0.0281&0.0300&0.0452&0.0434&0.0445&0.0441&0.0458&0.0473&\underline{0.0489}&\textbf{0.0555}&13.4\%\\	
&N@5&0.0358&0.0403&0.0600&0.0571&0.0584&0.0587&0.0591&0.0600&\underline{0.0624}&\textbf{0.0705}&12.9\%\\			

\hline
\multirow{4}{*}{Yelp}
&H@5&0.0202&0.0345&0.0359&0.0327&0.0351&0.0363&0.0368&\underline{0.0374}&\underline{0.0374}&\textbf{0.0420}&12.2\%\\
&P@5&0.0258&0.0374&0.0459&0.0418&0.0449&0.0465&0.0470&0.0478&
\underline{0.0479}&\textbf{0.0537}&12.1\%\\	
&R@5&0.0236&0.0386&0.0426&0.0365&0.0395&0.0419&0.0431&\underline{0.0437}&0.0432&\textbf{0.0500}&15.7\%\\	
&N@5&0.0315&0.0472&0.0580&0.0513&0.0551&0.0579&0.0603&0.0606&\underline{0.0607}&\textbf{0.0691}&13.8\%\\

\hline
\end{tabular}
}
\end{center}
\end{table*}

\begin{table*}[htbp]
\centering
\vspace{-2mm}
\caption{Overall Top-10 recommendation performance comparison of different recommender models.}
\vspace{-3mm}
\label{Table:overall1}
\renewcommand\arraystretch{1.0}
\begin{center}
\resizebox{0.93\linewidth}{!}{ 
\begin{tabular}{cc|cccccccccc|c}
\hline
Dataset&Metric&BPR&DiffNet&LightGCN&ESRF&SEPT&MHCN&DESIGN&DSL&SHaRe&IDVT&Improv.\\ \hline
\hline

\multirow{4}{*}{Flickr}
&H@10&0.0031&0.0038&0.0090&0.0054&0.0057&0.0083&0.0091&0.0082&\underline{0.0092}&\textbf{0.0101}&9.7\%\\
&P@10&0.0024&0.0029&0.0069&0.0042&0.0044&0.0063&\underline{0.0070}&0.0063&\underline{0.0070}&\textbf{0.0077}&10.0\%\\		
&R@10&0.0045&0.0047&0.0099&0.0058&0.0067&0.0112&\underline{0.0117}&0.0097&\underline{0.0117}&\textbf{0.0130}&11.1\%\\			
&N@10&0.0038&0.0044&0.0100&0.0061&0.0064&0.0097&\underline{0.0110}&0.0097&\underline{0.0110}&\textbf{0.0123}&11.8\%\\

\hline
\multirow{4}{*}{Ciao}	
&H@10&0.0425&0.0434&0.0670&0.0666&0.0679&0.0695&0.0678&0.0670&\underline{0.0709}&\textbf{0.0774}&9.1\%\\	
&P@10&0.0230&0.0278&0.0362&0.0360&0.0368&0.0376&0.0367&0.0362&\underline{0.0384}&\textbf{0.0419}&9.1\%\\	
&R@10&0.0460&0.0473&0.0708&0.0692&0.0695&0.0732&0.0727&0.0729&\underline{0.0780}&\textbf{0.0855}&9.6\%\\	
&N@10&0.0398&0.0441&0.0650&0.0621&0.0631&0.0651&0.0647&0.0650&\underline{0.0688}&\textbf{0.0761}&10.6\%\\		

\hline
\multirow{4}{*}{Yelp}
&H@10&0.0349&0.0564&0.0608&0.0557&0.0588&0.0615&0.0601&0.0610&\underline{0.0635}&\textbf{0.0688}&8.3\%\\	
&P@10&0.0223&0.0305&0.0389&0.0356&0.0376&0.0393&0.0384&0.0390&\underline{0.0406}&\textbf{0.0440}&8.3\%\\	
&R@10&0.0391&0.0596&0.0695&0.0602&0.0629&0.0668&0.0698&0.0688&\underline{0.0735}&\textbf{0.0792}&7.7\%\\	
&N@10&0.0351&0.0520&0.0639&0.0564&0.0595&0.0629&0.0655&0.0644&\underline{0.0676}&\textbf{0.0746}&10.3\%\\		
\hline
\end{tabular}
}
\end{center} 
\end{table*}

\begin{table*}[h]
\centering
\vspace{-2mm}
\caption{Ablation studies on different views.}
\vspace{-3mm}
\label{tab:ablation}
\renewcommand\arraystretch{1}
\begin{center}
\resizebox{0.93\linewidth}{!}{ 
\begin{tabular}{c|cccc|cccc|cccc}
\hline
\textbf{Dataset} & \multicolumn{4}{c|}{Flickr} & \multicolumn{4}{c|}{Ciao} & \multicolumn{4}{c}{Yelp} \\ \hline
Method      &H@5 &P@5 &R@5 &N@5                 &H@5 &P@5 &R@5 &N@5                 &H@5 &P@5 &R@5 &N@5          \\ \hline\hline
w/o both    &0.0054&0.0083&0.0073&0.0103        &0.0433&0.0469&0.0477&0.0618        &0.0374&0.0479&0.0441&0.0614   \\
w/o LV      &0.0055&0.0084&0.0075&0.0107        &0.0471&0.0511&0.0509&0.0669        &\underline{0.0416}&\underline{0.0532}&\underline{0.0497}&\underline{0.0687}   \\
w/o DV    &\underline{0.0059}&\underline{0.0090}&\underline{0.0079}&\underline{0.0112}        &\underline{0.0480}&\underline{0.0520}&\underline{0.0540}&\underline{0.0686}        &0.0404&0.0517&0.0485&0.0672   \\
IDVT    &\textbf{0.0062}&\textbf{0.0095}&\textbf{0.0083}&\textbf{0.0116}        &\textbf{0.0496}&\textbf{0.0537}&\textbf{0.0555}&\textbf{0.0705}                &\textbf{0.0420}&\textbf{0.0537}&\textbf{0.0500}&\textbf{0.0691}                                    \\  \hline
\end{tabular}
}
\end{center}
\end{table*}

\section{Experiments and Discussion}
\subsection{Experimental Setup}
\subsubsection{Datasets} 
We conduct experiments on three benchmark datasets: Flickr, Ciao, and Yelp. 
We first remove users without social connections and then apply a 5-core setting for the interaction data, retaining users and items with at least 5 interactions. The detailed data statistics are stated in Table \ref{tab:statistic2}. These three datasets have different levels of noise ratio, as demonstrated in Section III.A, which allows us to evaluate the robustness of our model in different scenarios.
\subsubsection{Metrics} 
To evaluate the top-$K$ recommendation performance of all methods, we employ four metrics: Hit Ratio (H), Precision (P), Recall (R), and Normalized discounted cumulative gain (N) with $K=5,10$. For each user in the test set, we conduct item ranking on all candidate items to ensure the reliability of the results.

\subsubsection{Comparison baselines} 
We compare IDVT with the following four categories of methods: traditional GNN-free, GNN-based collaborative filtering, GNN-based social, and SSL-based social models. In the first category, we include BPR\cite{rendle2012bpr}, which is a well-known method based on Bayesian personalized ranking. The second category consists of LightGCN\cite{he2020lightgcn} that focuses on collaborative filtering and explores the higher-order information in the interaction graph. In the third category, DiffNet\cite{wu2019neural} enhances user representations with recursive social diffusion, and DESIGN\cite{tao2022revisiting} uses the knowledge distillation technique to enhance the denoising process. In the fourth category, we consider ESRF \cite{yu2020enhancing}, SEPT \cite{yu2021socially}, MHCN\cite{yu2021self}, DSL\cite{wang2023denoised} and SHaRe~\cite{jiang2024challenging}, all capable of learning unlabelled data. 

\subsubsection{Implementation details} 
In our experiments, we fix the embedding size to 64 for all methods. The regulation coefficient $\lambda_3$ is set to 0.0001. We use a batch size of 2048 for efficient training and employ Adam optimization to update the model parameters during the learning process. For a comprehensive understanding of the model's performance, we provide detailed settings for other hyperparameters in the following paper. To make a fair comparison, all the compared experiments are conducted on a single Linux server with NVIDIA RTX A6000-48GB. Every compared algorithm was run 10 times for each reported result and the average value was taken as the final result.

\subsection{Overall Performance}
Table \ref{Table:overall} presents the comparative analysis of our model and the baselines, yielding the following observations: \textbf{(1)} Our model exhibits remarkable performance, with average improvements of 18\%, 14\%, and 13\% on the top 5 task, and 11\%, 10\%, and 9\% on the top 10 task, respectively, compared to the top-performing baseline across the three datasets. \textbf{(2)} The efficacy of GNN-based algorithms substantially surpasses that of conventional techniques like BPR. \textbf{(3)} Social recommendation models enriched with SSL exhibit superior performance compared to DiffNet, which lacks the capacity to learn from unlabelled data. \textbf{(4)} In the Flickr dataset characterized by a high noise ratio, LightGCN outperforms the self-supervised enhanced social recommendation models (ESRF, SEPT, and MHCN). Conversely, on the Ciao and Yelp datasets featuring low noise ratios, MHCN showcases superior performance over LightGCN. This observation suggests that social recommendation models lacking a denoising process do not achieve superior performance compared to methods that do not incorporate social connections when operating within environments with significant noise present in social networks. \textbf{(5)} Collectively, DESIGN demonstrates strong performance on the Flickr dataset characterized by a significant noise ratio, attributed to its employment of knowledge distillation to eliminate social network noise. Conversely, SHaRe exhibits superior performance on the Ciao and Yelp datasets, where the proportion of noise is comparatively smaller. SHaRe rewires social graph by cutting unreliable edges and adding highly homophilic edges.

\begin{figure*}[h]
\centering
\includegraphics[width=0.95\linewidth]{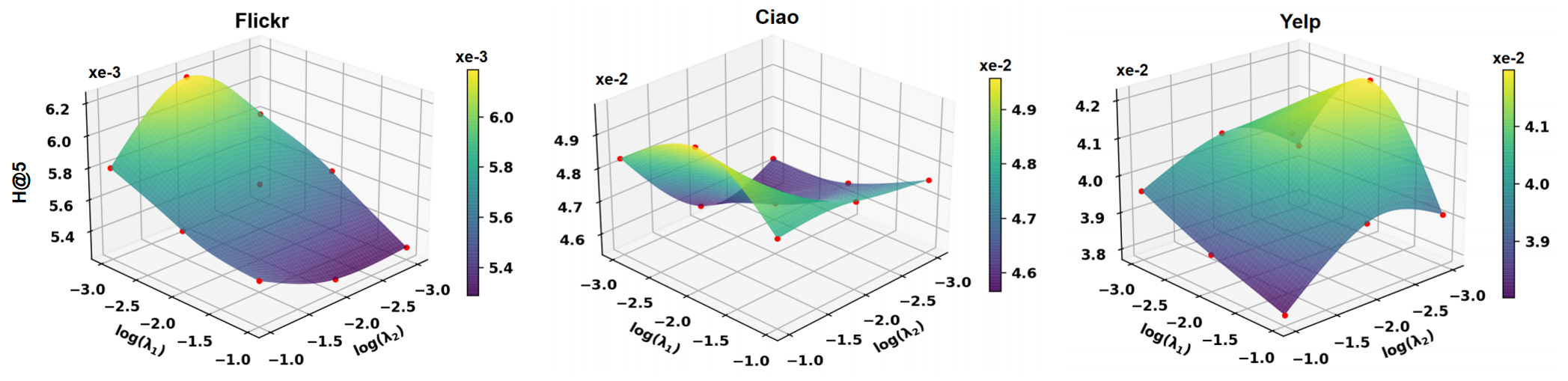} 
\caption{Sensitivity on weights of two view-guided tuning modules $\lambda_1$ and $\lambda_2$.} 
\vspace{-2mm}
\label{fig:lamda} 
\end{figure*}

\begin{figure*}[h]
\centering
\includegraphics[width=0.9\linewidth]{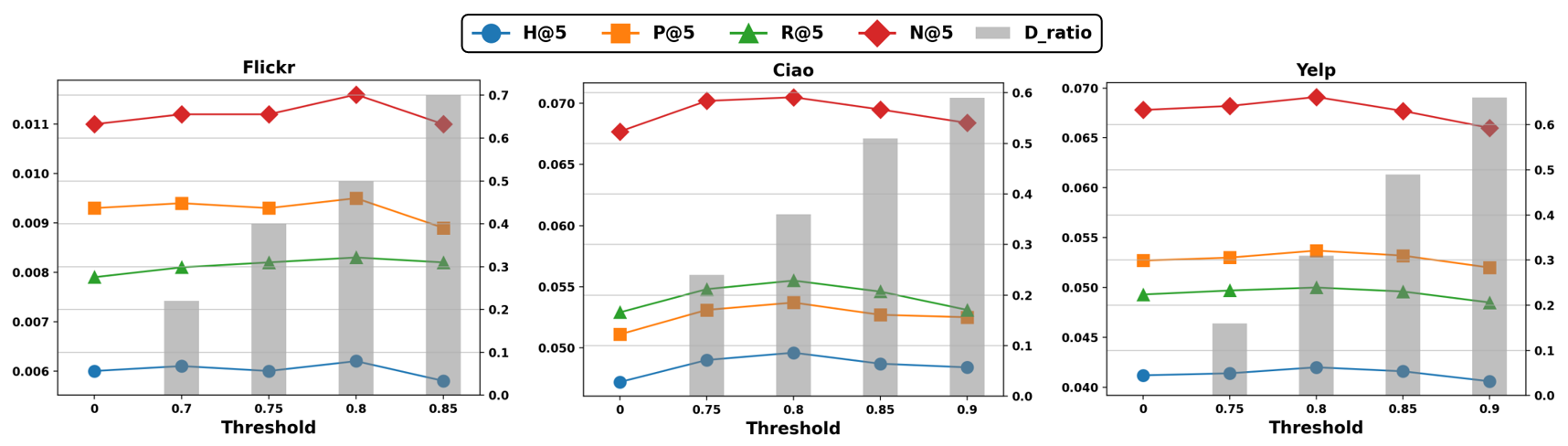} 
\vspace{-4mm}
\caption{Sensitivity on threshold of the noise removal ratio in the social graph.} 
\label{fig:threshold} 
\end{figure*}

\begin{figure*}[t]
\centering
\includegraphics[width=0.9\linewidth]{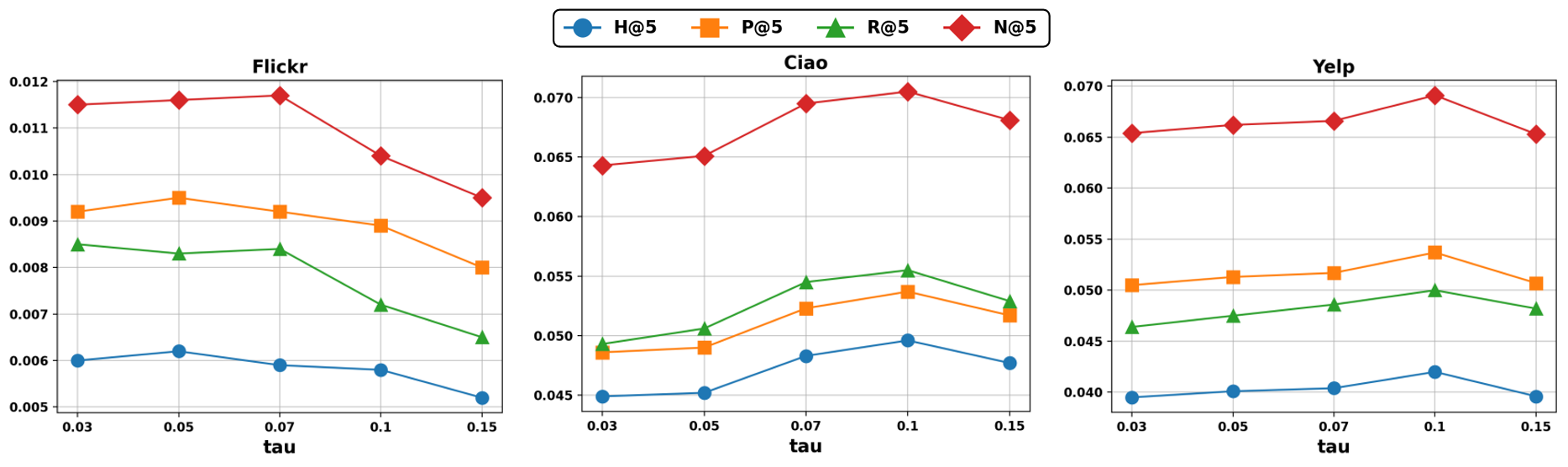} 
\vspace{-3mm}
\caption{Sensitivity on $\tau$ in the contrastive learning.} 
\label{fig:tau} 
\end{figure*}

\vspace{-3mm}
\subsection{Model Analyses}
\subsubsection{Ablation Study}
To assess the effectiveness of each module in fine-tuning the primary module, we conduct ablation experiments on three datasets. The results are presented in Table \ref{tab:ablation}, where ``w/o both", ``w/o LV", and ``w/o DV" denote the removal of both auxiliary views, removal of Local View, and removal of Dropout View, respectively. The experimental findings demonstrate that each view contributes to improving the model's performance. Specifically, on the Flickr and Ciao datasets, ``w/o DV" outperforms ``w/o LV", while on the Yelp dataset, the situation is reversed. These outcomes serve as a foundation for further validation in subsequent experiments.

\subsubsection{Sensitivity on $\lambda_1$ and $\lambda_2$}
The weights $\lambda_1$ and $\lambda_2$ are crucial in balancing the importance of two view-guided tuning modules. The sensitivity of these two parameters is depicted in Fig.\ref{fig:lamda}. Notably, on datasets with higher social noise ratios, such as Flickr and Ciao, the optimal combinations were found to be $\{\lambda_1=0.001, \lambda_2=0.01\}$ and $\{\lambda_1=0.01, \lambda_2=0.1\}$ respectively. In both cases, the value of $\lambda_2$ exceeded $\lambda_1$, indicating the importance of the Dropout View in fine-tuning denoising. Conversely, for the dataset with lower noise, Yelp, the best combination was $\{\lambda_1=0.01, \lambda_2=0.001\}$, suggesting the greater influence of the Local View in interest-aware fine-tuning. Remarkably, these experimental findings align with the ablation study, further supporting the efficacy and relevance of the proposed view-guided tuning approach.

\subsubsection{Sensitivity on Threshold} 
The threshold is a critical hyperparameter that controls the removal ratio of the social graph. We evaluate the model's performance by varying the threshold to five different values. The results are depicted in Fig.\ref{fig:threshold}. The line chart represents the values of the evaluation metrics, while the bar chart denotes the corresponding removing ratio under each threshold value. The experimental results indicate that for the three datasets with different social noise levels, the optimal removing ratios are found to be 50\%, 37\%, and 31\%, respectively. Impressively, these optimal ratios align well with the noise levels of the datasets, with higher noise datasets requiring a higher removal ratio. This consistency indicates the effectiveness of the model's performance and demonstrates its adaptability to datasets with diverse noise levels.

\subsubsection{Sensitivity on $\tau$} 
The hyperparameter $\tau$ plays a crucial role in facilitating hard negative mining, where ``hard negative" refers to nodes with similar representations. A smaller $\tau$ value increases the sensitivity of the hard negative samples. In our study, we experimented with five values within the interval [0.03, 0.15] to investigate their impact on the experimental results. Our findings indicate that both excessively large and small $\tau$ values lead to suboptimal performance. Remarkably, at the optimal performance, the $\tau$ value for the Flickr dataset is slightly lower than that of the other two datasets. This observation is attributed to the higher noise ratio in social networks present in the Flickr dataset. 

\begin{figure}[h]
\centering
\vspace{-2mm}
\includegraphics[width=0.75\columnwidth]{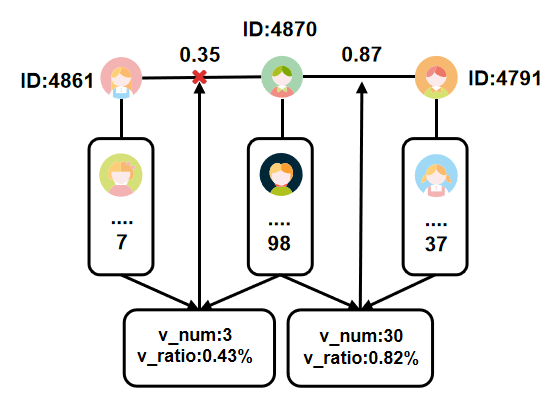}
\vspace{-3mm}
\caption{Case study on Interest-aware Denoising.}
\vspace{-5mm}
\label{fig:case_study} 
\end{figure}

\subsection{A Case Study on Interest-aware Denoising}
In this section, we present an example to validate the effectiveness of the Interest-aware Denoising module, as illustrated in Fig.\ref{fig:case_study}. To do so, we randomly select a user with ID:4870 and consider two pairs of social relationships, namely \{ID:4870, ID:4861\} and \{ID:4870, ID:4791\}. Among these pairs, one is removed, and the other one is retained. The three users involved in these pairs have 7, 98, and 37 social relationships, respectively. We define $v$\_num and $v$\_ratio to represent the number and ratio of shared interactions between the two groups of users. Remarkably, the $v$\_num and $v$\_ratio of \{ID:4870, ID:4861\} are significantly lower than those of \{ID:4870, ID:4791\}. These findings are further supported by the corresponding Interest Confidence (IC) values of 0.35 and 0.87, respectively, which perfectly align with our initial hypothesis. This example serves as a compelling demonstration of the Interest-aware Denoising module's ability to accurately denoise social connections and identify valuable and meaningful relationships in social networks for improved recommendation accuracy.

\begin{figure}[h]
\centering
\vspace{-2mm}
\includegraphics[width=1\columnwidth]{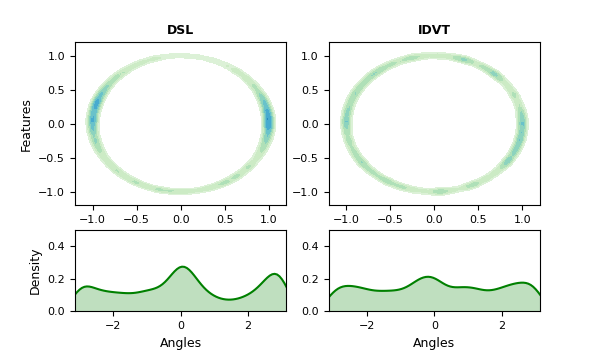} 
\vspace{-2mm}
\caption{Visualization of user embeddings on Flickr dataset.}
\label{fig:distribution} \vspace{-5mm}
\end{figure}

\subsection{Visualizing the Distribution of Representations}
To provide additional insight into the effectiveness of our model, we utilize Gaussian Kernel Density Estimation (KDE) to visualize the distribution of user representations in a two-dimensional space for both DSL and IDVT. As depicted in Fig.\ref{fig:distribution}, DSL displays distinct clustering of user representations, while the representations of IDVT appear more uniformly distributed. In alignment with SimGCL, a more uniform distribution of representations helps preserve the intrinsic characteristics of nodes and improves the generalization capabilities of the model.

\section{Conclusion}
In this paper, we conduct a comprehensive investigation into the noise present in social networks and its impact on recommendation systems. Based on the insights gained from data analysis, we propose a novel primary module that includes an innovative ID to denoise social connections. In this module, we also integrate both social and collaborative domains to improve user representations. Additionally, to tackle potential user interest loss and enhance model robustness in the primary module, we introduce two novel supplementary views, each in its respective auxiliary module, collectively forming a two-level VT process. Extensive experiments on three social network datasets demonstrate the flexibility and effectiveness of our proposed model. Ablation experiments are also conducted to elucidate the effects of the two supplementary views in the tuning process.

\bibliographystyle{unsrt} 
\bibliography{ICDM}

\end{document}